# Chief Complaint Classification with Recurrent Neural Networks

Scott Lee[1], Drew Levin[2], Pat Finley[2], Chad Heilig[1]

**Author Affiliations**
1 Centers for Disease Control and Prevention, Atlanta, GA
2 Sandia National Laboratories, Albuquerque, NM

**ABSTRACT**
Syndromic surveillance detects and monitors individual and population health indicators through sources such as emergency department records. Automated classification of these records can improve outbreak detection speed and diagnosis accuracy. Current syndromic systems rely on hand-coded keyword-based methods to parse written fields and may benefit from the use of modern supervised-learning classifier models. In this paper we implement two recurrent neural network models based on long short-term memory (LSTM) and gated recurrent unit (GRU) cells and compare them to two traditional bag-of-words classifiers: multinomial naïve Bayes (MNB) and a support vector machine (SVM). The MNB classifier is one of only two machine learning algorithms currently being used for syndromic surveillance. All four models are trained to predict diagnostic code groups as defined by Clinical Classification Software, first to predict from discharge diagnosis, then from chief complaint fields. The classifiers are trained on 3.6 million de-identified emergency department records from a single United States jurisdiction. We compare performance of these models primarily using the $F_1$ score. We measure absolute model performance to determine which conditions are the most amenable to surveillance based on chief complaint alone. Using discharge diagnoses The LSTM classifier performs best, though all models exhibit an $F_1$ score above 96.00. The GRU performs best on chief complaints ($F_1$=47.38), and MNB with bigrams performs worst ($F_1$=39.40). Certain syndrome types are easier to detect than others. For examples, chief complaints using the GRU model predicts alcohol-related disorders well ($F_1$=78.91) but predicts influenza poorly ($F_1$=14.80). In all instances the RNN models outperformed the bag-of-word classifiers suggesting deep learning models could substantially improve the automatic classification of unstructured text for syndromic surveillance.

**INTRODUCTION**
Syndromic surveillance—detection and monitoring individual and population health indicators that are discernible before confirmed diagnoses are made (Mandl et al.2004)—can draw from many data sources. Electronic health records of emergency department encounters, especially the free-text chief complaint field, are a common focus for syndromic surveillance (Yoon, Ising, & Gunn 2017). In practice, a computer algorithm associates the text of the chief complaint field with predefined syndromes, often by picking out keywords or parts of keywords or a machine learning algorithm based on mathematical representation of the chief complaint text. In this paper, we explore recurrent neural networks as an alternative to existing methods for associating chief complaint text with syndromes.

**Overview of Chief Complaint Classifiers**
In a recent overview of chief complaint classifiers (Conway et al., 2013), the authors divide chief complaint classifiers into 3 categories: keyword-based classifiers, linguistic classifiers, and statistical classifiers.

Keyword-based classifiers associate a chief complaint text with a specific syndrome according to the presence of specific words, parts of words, or other character strings. Similarly, linguistic classifiers use a complex hierarchy of predefined rules to bin chief complaint records into syndrome categories. In contrast to keyword-based classifiers, linguistic classifiers try to exploit additional information about the lexicon and variations in word formation to accomplish their task. Algorithms that explicitly account for negation, like NegEx (Chapman 2001), may serve as components in linguistic systems, as may algorithms that use contextual information to determine whether mentions of a condition, like fever or vomiting, indicate its actual presence (Chapman 2007). Statistical chief complaint classifiers predict syndromes based on models learned from examples that have been previously labeled.

Modern statistical classifiers can capture the decision rules employed by many keyword-based classifiers. For example, the classifier deployed in the Electronic Surveillance System for the Early Notification of Community-Based Epidemics (ESSENCE) is keyword-based (Conway et al. 2013). Under this method, the record and the syndrome are represented as an unordered set of words or tokens, with a numeric value or weight assigned to each token. When the sum of weights exceeds a predetermined threshold, the record is flagged as an example of the syndrome. This method effectively constructs a linear classifier, but it is one where both the weights (i.e., the syndrome vector) and the decision boundary (i.e., the threshold) are manually determined rather than learned. A machine-learned linear classifier like a support vector machine (SVM) could capture this rule precisely, but it could also capture the expert judgement that informed the rule in the first place. Thus, many keyword-based classifiers may be considered a special, hand-tuned kind of statistical classifier.

Furthermore, statistical classifiers can often mimic the performance of linguistic classifiers. For example, deep recurrent neural networks (RNNs) are capable of learning rich representations of words, sentences, and even entire documents from sequences of text. Machine translation, which for decades was dominated by rule-based, hand-engineered systems, has been largely replaced by RNNs; these models are trained with only natural language as input, and they have achieved remarkable results on difficult tasks, including zero-shot translation, i.e. translating between language pairs not explicitly seen during training (Johnson et al 2016). Going beyond RNNs, even models invented primarily for non-linguistic tasks, like image classification, have been used to solve complex problems with text, like understanding the sentiment behind user-generated restaurant ratings and product reviews (Kim 2014; Blunsom 2014; Zhang 2015).

Since chief complaint classification is a kind of text categorization, methods good for the latter serve the former as well. Thus, we compare several popular, open-source, supervised learning algorithms for classifying chief complaints, in contrast to proprietary systems that have been developed specifically for syndromic surveillance.

**Overview of Current Methods for Text Categorization**
Current methods for text categorization are generally non-sequential models, like naïve Bayes and SVMs, or sequential models, like one-dimensional convolutional neural networks (CNNs) and RNNs. Non-sequential models and CNNs have achieved state-of-the-art results on many benchmark datasets, including movie reviews (Wang 2012), product reviews (Zhang 2015), and news articles (Johnson 2014). Likewise, RNN models, especially long short-term memory (LSTM) neural networks (Hochreiter 1997) and their variants, have been used to great success on language modeling (Sundermeyer 2012), natural language understanding (Bowman 2015), and machine translation tasks (Bahdanau 2014; Wu et al. 2016), in addition to sentiment analysis and other kinds of text categorization (Mesnil 2014).

RNNs have only recently come into widespread use, where the structural characteristics of the LSTM and its variants, like the gated recurrent unit (GRU) (Cho et al. 2014), have allowed them to handle long-range dependencies in sequential data without suffering from the problem of vanishing or exploding gradients. Several studies have compared the performance of LSTM and GRU architectures (Chung 2015). Each has at times outperformed the other and can be considered comparable (Greff et al. 2017). When nested hierarchically, these models are also capable of classifying longer sequences of text (Tang et al. 2015; Yang et al. 2016). Jagannatha et al. (2016) showed that LSTMs substantially outperform conditional random fields (CRFs) for medical event detection in electronic health records, and Helwe et al. (2017) showed they could automatically label discharge diagnosis notes with Clinical Classification Software (CCS) (HCUP) categories with high accuracy. Their best-performing model achieved micro-averaged $F_1$ scores of well over 0.85 for the raw text and 0.95 for text that had been mapped to categories in a preexisting medical ontology.

**Goals of the Current Study**
In the current study, we examine four machine learning models applied to chief complaint text: multinomial naive Bayes (MNB) and support vector machines (SVM), as well as LSTM and GRU RNNs. The MNB classifier is one of only two machine learning algorithms currently being used for syndromic surveillance (Conway 2013), the other being a maximum entropy classifier (O'Connor 2007). Our first goal is to predict CCS codes from discharge diagnosis, as in Helwe et al. (2017). Medical coding is time-consuming, and accurate automatic classification could be useful. When applied to syndromic surveillance systems, improved coding speed could increase the speed of response to outbreaks, emerging infections, and other public health emergencies. Our second goal is to predict CCS codes from chief complaint, as the chief complaint is generally less informative than discharge diagnosis with respect to clinical outcomes but more useful for real-time surveillance since chief complaints are available much sooner, typically immediately after emergency department triage. We are interested in the performance of these models as compared to each other, as well as their absolute performance in predicting specific CCS codes, which could help practitioners determine which conditions are the most amenable to surveillance based on chief complaint alone. Following our comparison of these models, we briefly discuss the need for labeled data to train machine learning models.

**METHODS**
**Corpus and Data Structure**
Our original dataset comprised 3.6 million de-identified, patient-level ED visit records collected by the New York City Department of Health and Mental Hygiene in 2016. The data included several demographic variables, like sex and age group, as well as date and time of admission and a hospital code. Our focus for this analysis was exclusively on three text fields: chief complaint (CC), discharge diagnosis (DD), and ICD-9/10 diagnosis code. The analysis includes only the 2.1 million emergency department records that contain a nonempty CC or DD field and a nonempty ICD field. The remaining 1.5 million records in the original data were missing at least one of these and were thus omitted from further analysis.

Although the content varies from hospital to hospital, a typical record will have a short CC, like "fever and cough"; a longer, more technical DD, like "suppurative otitis media in diseases classified elsewhere"; and at least one ICD diagnosis code, like 382.02. As with this example, many of the DD descriptions in our dataset were automatically generated during the coding process, lowering the variability in their mapping to the CCS codes, and making them much easier to classify with a learned model. Some of the descriptions were not drawn directly from the ICD dictionary, however, like "polysubstance dependence" for the ICD-10 code F19.20, which is fully defined as "other psychoactive substance dependence, unspecified", and the descriptions often overlap in terms of their vocabulary, so the classification task is not entirely trivial.

We preprocessed the CC and DD text fields by removing special characters, converting digits to their written forms (e.g., '4' to 'four'), and then lowercasing the remaining text. For this study, we omitted additional steps, like expanding abbreviations, correcting misspellings, or mapping words to concepts in a medical ontology like the Unified Medical Language System (UMLS). We also preprocessed the ICD codes, which were stored as pipe-delimited text strings in the original dataset. First, we removed special characters from the codes, including the pipe delimiters and punctuation, and then we converted them to lowercase. All ICD-9 and ICD-10 codes were replaced by their corresponding CCS codes. Duplicate CCS codes in a single record were removed. Records with more than one unique CCS code were omitted from this analysis.

For our MNB and SVM classifiers, we represent each record as a single document in a bag-of-words (BoW) model. Under this model, each document $d$ is represented as a row vector of word counts, where each entry in the row corresponds to the number of times a particular word $w$ appears in the document. The entire corpus is represented as a $D$ x $V$ document-term matrix, where $D$ is the number of records in the corpus, and $V$ is the number of unique tokens in the corpus, or vocabulary. Each row, then, is the BoW vector for a particular CC or DD record. To make our classifiers more effective, we count both single words, or unigrams, and pairs of adjacent words, or bigrams. We did not use term frequency inverse document frequency (TF-IDF) transformations to remain consistent with known syndromic surveillance implementations.

For our LSTM and GRU RNN models, we represent each word as a one-hot column vector, where each entry is 0 except for the one corresponding to the word's index in the vocabulary; this index is set to 1. We represent a single record as a sequence of such vectors, padding the sequence with column vectors of zeros so that the number of columns in the resulting matrix matches the number of words in the longest record in the corpus. The full collection of records, then, is represented as a $V$ x $L_{max}$ x $D$ tensor, where $L_{max}$ is length in words of the longest record, and $V$ and $D$ are the same as above. This data structure preserves the sequential information in the records, and it allows the first layer of the network to function as a lookup matrix for the word embeddings, which are learned with the rest of the model weights during training. To use CCS codes as targets for the RNNs, we converted the codes from strings to integers and then encoded them as one-hot row vectors.

Because de-identified data do not constitute human subjects, this analysis did not require review by an institutional review board.

**Model Structure and Implementation**

We implemented our MNB classifier with uniform priors and the Laplace/Lidstone smoothing parameter set to 1.0; these are the default parameters in scikit-learn, and adjusting them did not yield substantial increases in performance. Similarly, we implemented our SVM using the library's default parameters, including a linear kernel, L2 penalty, square hinge loss, and a one-vs-rest classification scheme to handle our multiclass outputs. Both models took the bag-of-bigrams document-term matrix as their input and the column vector of single CCS codes as their output.

We implemented both our LSTM and GRU as bidirectional RNNs (Graves 2005) with a 200-dimensional embedding layer and a 100-dimensional hidden layer. Unidirectional versions of both models performed worse. To reduce overfitting, we applied dropout with a probability of 0.5 to the concatenated outputs of the forward and backward cells, and we used a dense layer to obtain predicted class probabilities via softmax. We measured the fit of the model with categorical cross-entropy loss, and we used the Adam algorithm with a learning rate of 0.001 (Kingma and Ba 2014) for optimization.

We limited our analysis to models that predict only one classification at a time to facilitate direct comparisons among our 4 models. Although an RNN model can predict multiple labels through a change to the loss function, this extension exceeds the scope of the current analysis.

The MNB classifier and SVM were built using the scikit-learn 0.19.1 Python library (Pedregosa et al. 2011), and the LSTM and GRU were built using Keras 2.3.1 (Chollet 2015) with the TensorFlow (Abadi et al. 2016) backend.

**Experimental Setup**

We assessed the performance of all models using a single random training-validtion-test split, using 60% of the records for training, 20% for model tuning, and the final 20% for testing. We trained the LSTM and GRU in batches of 256 records for a maximum of 10 epochs, stopping after 2 epochs if the validation loss did not decline and saving the model with the lowest validation loss for testing.

To evaluate the models, we calculated their sensitivity (also known as recall), positive predictive value (PPV, also known as precision), and $F_1$ score (the harmonic average of precision and recall) in predicting each of the unique CCS codes in the test data. To adjust for class imbalance, we used a weighted macro-average of these individual scores to serve as measures of overall classification performance, and we used statistical methods for paired data to compare models based on their code-specific classification results. For these results, we also present the predicted prevalence per 1,000 ED visit records of each condition in the test set. In clinical settings, scores like PPV and $F_1$ are important, since models are responsible for making accurate individual-level predictions. In surveillance settings, the predicted proportion with the conditions is also important, since the number of positive calls a model makes is used to detect outbreaks and forecast disease trends. Models with low classification accuracy do not necessarily produce poor prevalence estimates.

## RESULTS

In our corpus, the CC records had more unique tokens and greater average length than the DD records (Table 1), in part because the CC text often includes abbreviations, acronyms, misspellings, and (expanded) diagnosis codes.

|    | Total records | Total tokens | Unique tokens | Min, med, max (IQR) | Mean (SD) |
|----|---------------|--------------|---------------|---------------------|-----------|
| DD | 2,061,134     | 11,072,256   | 20,210        | 1, 4, 46 (*3-7*)    | 5.37 (4.09) |
| CC | 2,046,616     | 15,796,329   | 74,489        | 1, 4, 82 (*2-10*)   | 7.72 (9.12) |

Table 1. Descriptive statistics for the discharge diagnosis (DD) and chief complaint (CC) fields in our dataset, including the minimum (min), median (med), and maximum (max) number of tokens per record.

### Predicting CCS Code from Discharge Diagnosis

As we expected, all 4 models were able to predict CCS codes from the DD text with high accuracy (Table 2). The worst-performing model, the MNB classifier, still achieved an $F_1$ score of over 0.96. The GRU posted the best performance, with all scores over 99.60, and the LSTM was very close behind.

Given the similarity between LSTM and GRU scores, we expect that they perform close to the Bayes error rate (BER) for the test data. The GRU's performance on specific codes was equally impressive, reaching a perfect $F_1$ of 1.0 on 48 of the 282 CCS codes, and predicting equal prevalences for 86 codes. By contrast, the MNB classifier achieved an $F_1$ of 1.0 on only 1 code and predicted equal prevalences for only 7 codes; the SVM fared better with perfect scores on 21 codes and equal prevalences for 33 codes.

| Model        | Sens  | PPV   | $F_1$ |
|--------------|-------|-------|-------|
| $MNB_{uni}$  | 96.45 | 96.31 | 96.20 |
| $MNB_{bi}$   | 97.50 | 97.46 | 97.32 |
| $SVM_{uni}$  | 97.80 | 97.82 | 97.76 |
| $SVM_{bi}$   | 98.88 | 98.88 | 98.86 |
| LSTM         | 99.56 | 99.56 | 99.57 |
| **GRU**      | **99.65** | **99.65** | **99.65** |

Table 2. Weighted sensitivity (Sens), positive predictive value (PPV), and $F_1$ scores for each of our models on the discharge diagnosis (DD) classification task. Subscript "uni" indicates models with unigrams only; subscript "bi" indicates models with both unigrams and bigrams.

### Predicting CCS Code from Chief Complaint

The scores for the chief complaint classification task are much lower than those for DD classification (Table 3). RNNs still had the best performance, with the GRU achieving a weighted $F_1$ score of 47.38, and the LSTM at 47.30. The MNB classifier fared worse overall with bigram features than with unigram features, with $F_1$ scores of 39.4 and 40.53, respectively. The SVM did not show this pattern, with the bigram model scoring about a percentage point higher on $F_1$ than the unigram model.

| Model | Sens | PPV | $F_1$ |
|---|---|---|---|
| $MNB_{uni}$ | 44.72 | 47.99 | 40.53 |
| $MNB_{bi}$ | 43.90 | 51.30 | 39.40 |
| $SVM_{uni}$ | 42.33 | 42.08 | 39.84 |
| $SVM_{bi}$ | 45.75 | 45.40 | 42.82 |
| LSTM | **49.83** | 50.23 | 47.30 |
| GRU | 49.77 | **50.77** | **47.38** |

Table 3. Weighted macro sensitivity (Sens), positive predictive value (PPV), and $F_1$ scores for each of our models on the chief complaint (CC) classification task. Subscript "uni" indicates models with unigrams only; subscript "bi" indicates models with both unigrams and bigrams.

Table 4 shows $F_1$ scores for each classifier on a selection conditions commonly tracked by syndromic surveillance system. As indicated by the macro scores in Table 3, the LSTM and GRU generally outperform the other models, although the SVM posts the best score for influenza. This trend holds for the full range of codes, as well, with the RNNs achieving the highest scores for 190 of the 282 conditions in the test data. By contrast, the SVMs achieve the highest score on only 21 conditions, and the MNB on only 9 (see the supplemental materials for the full table of results). Although the difference in accuracy between the GRU and the best-performing MNB classifier is only about 5% (95% CI 4.94-5.17), the GRU achieves a higher $F_1$ on the majority of conditions. Finally, we note that despite the similarity of their macro scores, the GRU performs significantly better by this measure than the LSTM (Wilcoxon signed-rank test; p=0.001).

Because it is our best-performing model, we further explore the performance of GRU models for CC classification in Table 5. Classification accuracy was the highest for codes corresponding to conditions that are easier for clinicians to identify during the triage process, like burns ($F_1$=83.86), epilepsy and convulsions ($F_1$=80.82), disorders of the teeth and jaw ($F_1$=81.96), and cardiac arrest ($F_1$=72.94). GRU performed well with some syndromic conditions, such as alcohol-related disorders ($F_1$=78.91), while viral infection ($F_1$=25.68) and influenza ($F_1$=14.80) were much more difficult to classify from chief complaint. Generally, conditions with high $F_1$ scores also had a low difference between true and predicted prevalence, but so did a number of conditions with low $F_1$ scores, like other lower respiratory disease ($F_1$=35.57; rel.diff.=8.87%). Although the 4 model structures exhibit variation in predictive accuracy, they tend to produce a similar rank-order of the 144 CCS codes that appear in all of their predictions on the test data, with Spearman rank correlation of 0.9111 and Kendall's tau of 0.7701. Thus, the CCS codes for which the GRU tends to perform best are codes for which the other models also tend to perform best.

| CCS | Description | GRU | LSTM | MNB$_{bi}$ | MNB$_{uni}$ | SVM$_{bi}$ | SVM$_{uni}$ |
| --- | --- | --- | --- | --- | --- | --- | --- |
| 660 | Alcohol-related disorders | 78.91 | **79.10** | 69.58 | 72.68 | 74.41 | 65.45 |
| 128 | Asthma | 68.39 | **68.49** | 65.61 | 64.05 | 64.54 | 63.15 |
| 251 | Abdominal pain | 53.64 | **53.98** | 44.39 | 48.72 | 51.70 | 42.01 |
| 134 | Other upper respiratory disease | 51.72 | **51.99** | 42.50 | 45.73 | 45.44 | 45.22 |
| 250 | Nausea and vomiting | **41.81** | 41.76 | 19.31 | 33.74 | 36.39 | 26.71 |
| 133 | Other lower respiratory disease | **35.57** | 37.38 | 29.71 | 29.27 | 22.53 | 26.17 |
| 7 | Viral infection | 25.68 | **33.30** | 21.41 | 23.53 | 22.67 | 11.48 |
| 242 | Poisoning by other medications and drugs | 22.18 | **25.50** | 15.76 | 16.00 | 22.82 | 19.78 |
| 123 | Influenza | 14.80 | 13.41 | 13.11 | 13.27 | **15.37** | 15.18 |

Table 4. F$_1$ scores of each model for select conditions, with the highest score in bold.

**DISCUSSION**

Among the models that we evaluated, the MNB classifier had the poorest performance, yet its design is the most similar to one of the chief complaint classifiers currently used in practice. The RNN models—the LSTM and GRU—performed substantially better than the BoW models. This difference was especially apparent on the discharge diagnosis task, where the MNB classifier achieved an F$_1$ score of only 0.9755 despite the majority of the input text being drawn verbatim from the ICD-9 and ICD-10 dictionaries. Although this result has little practical importance, it demonstrates the empirical benefits of modeling both semantic and sequential information in the text, which the RNNs do explicitly but the BoW models only can only approximate.

Even with the increased representational capacity afforded by the RNNs, chief complaint classification was still a difficult task. This might depend in part on the dimensionality of our output space; our set of approximately 280 target CCS codes is much larger than that of a typical syndrome categorization scheme, which may only include 10 or 20 codes. In addition, the chief complaint itself is preliminary, as it tries to assess a patient's condition before full clinical examination. Some conditions, like generic viral infection, were particularly difficult for our models to classify as evidence by the best F$_1$ score of 27.06 by the GRU. This might be explained by symptoms, like fever and cough, also appearing as symptoms of many other conditions reported in the ED. On the other hand, conditions with overt, non-overlapping symptoms, like burns, were relatively easy to classify, demonstrating that chief complaint is more useful for monitoring some conditions than others. Triage notes, which typically contain more information than chief complaint, could help close this gap, but they are often omitted from the hospital data feeds provided to health jurisdictions for monitoring.

| CCS | Description | True Per1k | Pred Per1k | Abs. Diff. | % Rel. Diff. | Sens | PPV | $F_1$ |
|---|---|---|---|---|---|---|---|---|
| 660 | Alcohol-related disorders | 37.46 | 39.86 | 2.39 | 6.39 | 81.43 | 76.54 | 78.91 |
| 128 | Asthma | 30.07 | 35.74 | 5.67 | 18.85 | 74.83 | 62.97 | 68.39 |
| 251 | Abdominal pain | 48.10 | 86.93 | 38.83 | 80.74 | 75.30 | 41.66 | 53.64 |
| 134 | Other upper respiratory disease | 42.49 | 62.95 | 20.46 | 48.14 | 65.79 | 44.41 | 53.03 |
| 250 | Nausea and vomiting | 11.99 | 16.71 | 4.708 | 39.24 | 50.01 | 35.92 | 41.81 |
| 133 | Other lower respiratory disease | 26.74 | 29.11 | 2.37 | 8.87 | 37.15 | 34.12 | 35.57 |
| 7 | Viral infection | 23.68 | 12.14 | -11.54 | 48.74 | 19.42 | 37.88 | 25.68 |
| 242 | Poisoning by other medications and drugs | 1.08 | 0.29 | -0.79 | 73.24 | 14.06 | 52.54 | 22.18 |
| 123 | Influenza | 2.02 | 0.28 | -1.79 | 88.74 | 8.23 | 73.12 | 14.80 |

Table 4. Performance of a bidirectional gated recurrent unit (GRU) in predicting selected Clinical Classification Software (CCS) codes from chief complaint. Metrics shown are predicted and true prevalence per 1,000 ED visits; absolute difference between the prevalences; relative difference between the prevalences, as defined by |predicted prevalence - true prevalence| / true prevalence; and 3 binary measures of diagnostic accuracy. CCS codes are listed in descending order of the $F_1$ score.

This analysis assumed that the ICD diagnosis codes in the ED record, and in turn the corresponding unique CCS codes, could effectively serve as target proxies for true underlying conditions or syndromes. Our analysis could be extended in at least 2 ways. First, if we were able to use ED records that were labeled by experts and used to develop supervised machine learning models that predict those expert targets, then our methods could influence the practice of syndromic surveillance more directly. The main limitation here is labeling data, which is time-consuming and difficult to accomplish at scale because the sensitive nature of ED visit records prevents them from being easily shared outside of the jurisdictions where they are collected. However, once acquired, the data themselves would be relatively easy to categorize. Consider broad and intuitive examples, such as those where the North Carolina Disease Event Tracking and Epidemiologic Tracking Tool (NC DETECT) monitors syndromes categorized as "drowning", "trauma", "animal", and "assault". It could be feasible to obtain independently assessed labels for such syndromes and to create a large-scale labeled dataset to support research and to benchmark definitions against a reference standard. Modeling individual labelers (Guan et al. 2017) could also improve the overall accuracy of chief complaint classification algorithms, and resulting trained models could be shared with the data-providing jurisdictions to enhance their own surveillance systems. Unsupervised methods, like the representation learning algorithms presented in Y. Choi (2016) and E. Choi (2016), can capture some information in syndromic records, but supervised methods are likely the key to improving the robustness and scalability of our existing surveillance systems, and so we believe labeling to be a worthwhile pursuit.

Second, our analysis might be extended by including records that map to more than one CCS code. We simplified our approach by focusing on records that map to a unique CCS code. Our methods, however, could extend to classifying multiple syndromes associated with a single chief complaint text. All 4 models could be trained on multiple labels (that is, CCS codes) as a set of separate models 1 discrete, binary label per model. In contrast, a single RNN model can be trained to predict multiple labels simultaneously, for example by using binary instead of categorical cross-entropy loss to train our RNNs. The latter approach, although elegant, has the disadvantage of needing to retrain the whole model as new targets are considered, while the former approach can be implemented in a modular, more flexible way.

**CONCLUSION**

The RNNs outperformed the MNB classifier and SVM on all classification tasks. We conclude that deep learning models could substantially improve the automatic classification of unstructured text for syndromic surveillance.

**ACKNOWLEDGEMENTS**

We would like to thank Ramona Lall, Robert Mathes, and the BCD Syndromic Surveillance Unit at the New York City Department of Health and Mental Hygiene for providing the data for this study and for giving us valuable feedback during the early stages of analysis.

DL and PF acknowledge funding from the Sandia National Laboratory's Laboratory Directed Research and Development program. Sandia National Laboratories is a multimission laboratory managed and operated by National Technology & Engineering Solutions of Sandia, LLC, a wholly owned subsidiary of Honeywell International Inc., for the U.S. Department of Energy's National Nuclear Security Administration under contract DE-NA0003525.This paper describes objective technical results and analysis. Any subjective views or opinions that might be expressed in the paper do not necessarily represent the views of the U.S. Department of Energy or the United States Government.